\documentclass{article}

\usepackage{arxiv}

\usepackage[utf8]{inputenc} 
\usepackage[T1]{fontenc}    
\usepackage{hyperref}       
\usepackage{url}            
\usepackage{booktabs}       
\usepackage{amsfonts}       
\usepackage{nicefrac}       
\usepackage{microtype}      
\usepackage{lipsum}
\usepackage{graphicx}
\usepackage[utf8]{inputenc} 
\usepackage[T1]{fontenc}    
\usepackage{hyperref}       
\usepackage{url}            
\usepackage{booktabs}       
\usepackage{amsfonts}       
\usepackage{nicefrac}       
\usepackage{microtype}      
\usepackage{xcolor}         
\usepackage{graphicx}
\usepackage{subfigure}
\usepackage{amsmath}
\usepackage{amsthm}
\usepackage{amssymb}
\usepackage{enumitem}
\usepackage{adjustbox}
\usepackage{wrapfig}
\usepackage[normalem]{ulem}
\usepackage{mathrsfs}
\usepackage{multirow}
\usepackage{svg}

\usepackage{algorithm}
\usepackage{algorithmic}
\theoremstyle{definition}

\newcommand\algorithmicprocedure{\textbf{function}}
\newcommand{\algorithmicendprocedure}{\algorithmicend\ \algorithmicprocedure}
\makeatletter
\newcommand\FUNCTION[3][default]{%
  \ALC@it
  \algorithmicprocedure\ \textsc{#2}(#3)%
  \ALC@com{#1}%
  \begin{ALC@prc}%
}
\newcommand\ENDFUNCTION{%
  \end{ALC@prc}%
  \ifthenelse{\boolean{ALC@noend}}{}{%
    \ALC@it\algorithmicendprocedure
  }%
}
\newenvironment{ALC@prc}{\begin{ALC@g}}{\end{ALC@g}}

\graphicspath{ {./figures/} }

\title{General-Purpose Multi-Modal OOD Detection Framework}

\author{
  Viet Duong \\
  Department of Computer Science \\
  College of William and Mary \\
  \texttt{vqduong@wm.edu} \\
   \And
  Qiong Wu \\
  AT \&T Labs \\
  \texttt{qwu05@email.wm.edu} \\
  \And
  Zhengyi Zhou \\
  AT \&T Labs \\
  \texttt{zzhou@research.att.com} \\
  \And \And
  Eric Zavesky \\
  AT \&T Labs \\
  \texttt{ez2685@att.com} \\
  \And
  Jiahe Chen \\
  College of Control Science and Engineering \\
  Zhejiang University \\
  \texttt{mulplue@zju.edu.cn} \\
  \And
  Xiangzhou Liu \\
  School of Computer Science \\
  Zhejiang University \\
  \texttt{lyonl8639@gmail.com} \\
  \And
  Wen-Ling Hsu \\
  AT \&T Labs \\
  \texttt{wenlhsu@gmail.com} \\
  \And
  Huajie Shao \\
  Department of Computer Science \\
  College of William and Mary \\
  \texttt{hshao@wm.edu} \\
}

\begin{document}
\maketitle
\begin{abstract}
Out-of-distribution (OOD) detection identifies test samples that differ from the training data, which is critical to ensuring the safety and reliability of machine learning (ML) systems. While a plethora of methods have been developed to detect uni-modal OOD samples, only a few have focused on multi-modal OOD detection. Current contrastive learning-based methods primarily study multi-modal OOD detection in a scenario where both a given image and its corresponding textual description come from a new domain. However, real-world deployments of ML systems may face more anomaly scenarios caused by multiple factors like sensor faults, bad weather, and environmental changes. Hence, the goal of this work is to simultaneously detect from multiple different OOD scenarios in a fine-grained manner. To reach this goal, we propose a general-purpose weakly-supervised OOD detection framework, called WOOD, that combines a binary classifier and a contrastive learning component to reap the benefits of both. In order to better distinguish the latent representations of in-distribution (ID) and OOD samples, we adopt the Hinge loss to constrain their similarity. Furthermore, we develop a new scoring metric to integrate the prediction results from both the binary classifier and contrastive learning for identifying OOD samples. We evaluate the proposed WOOD model on multiple real-world datasets, and the experimental results demonstrate that the WOOD model outperforms the state-of-the-art methods for multi-modal OOD detection. Importantly, our approach is able to achieve high accuracy in OOD detection in three different OOD scenarios simultaneously. The source code will be made publicly available upon publication.

\end{abstract}


\section{Introduction}\label{sec:intro}
Out-of-distribution (OOD) detection~\cite{yang2021generalized,bogdoll2022anomaly,ming2022poem,ruff2021unifying,ma2021comprehensive} aims at identifying whether a test sample differs from the training data. Such detection is crucial for ensuring the safety and reliability of machine learning (ML) systems~\cite{wang2020safety,hussain2018autonomous}, such like autonomous driving and AI diagnosis~\cite{davenport2019potential,han2020bridging,macdonald2022interpretable}. For instance, autonomous driving systems should have the ability to detect unknown or unusual scenes in the real-world and issue an early warning to the driver to take control of the vehicle in a timely fashion to avoid potentially fatal accidents. Recent studies have developed many OOD detection models to enhance the safety and reliability of such ML systems.

Most existing studies focus on single-modal OOD detection, while multi-modal OOD detection is less explored. In real-world applications like autonomous driving, various types of sensors, including cameras, LiDAR, and radar, are used to enhance detection accuracy and enable safe and reliable decision-making. ML systems deployed in an open world, such as autonomous driving, also often encounter different anomaly scenarios caused by multiple factors, including sensor faults, bad weather, and environmental changes. However, recent works~\cite{mingdelving,esmaeilpour2022zero} on multi-modal OOD detection primarily examine one scenario where both an image and its textual description come from a new domain. Specifically, existing studies leverage CLIP-based techniques~\cite{ming2022delving,esmaeilpour2022zero,fort2021exploring} to identify whether a query image matches one of the labels (e.g., road) in the training data. However, these approaches can only detect \textit{visual OODs} rather than those arising from both the image and its corresponding textual information. As a result, they are not applicable to detect anomaly scenarios such as incorrect pairings of an image and its \textit{detailed textual description} (e.g., road is covered with snow).

\begin{figure*}[!tb]
\centering
\subfigure[Contrastive Learning Score.]{
\includegraphics[width=0.32\linewidth]{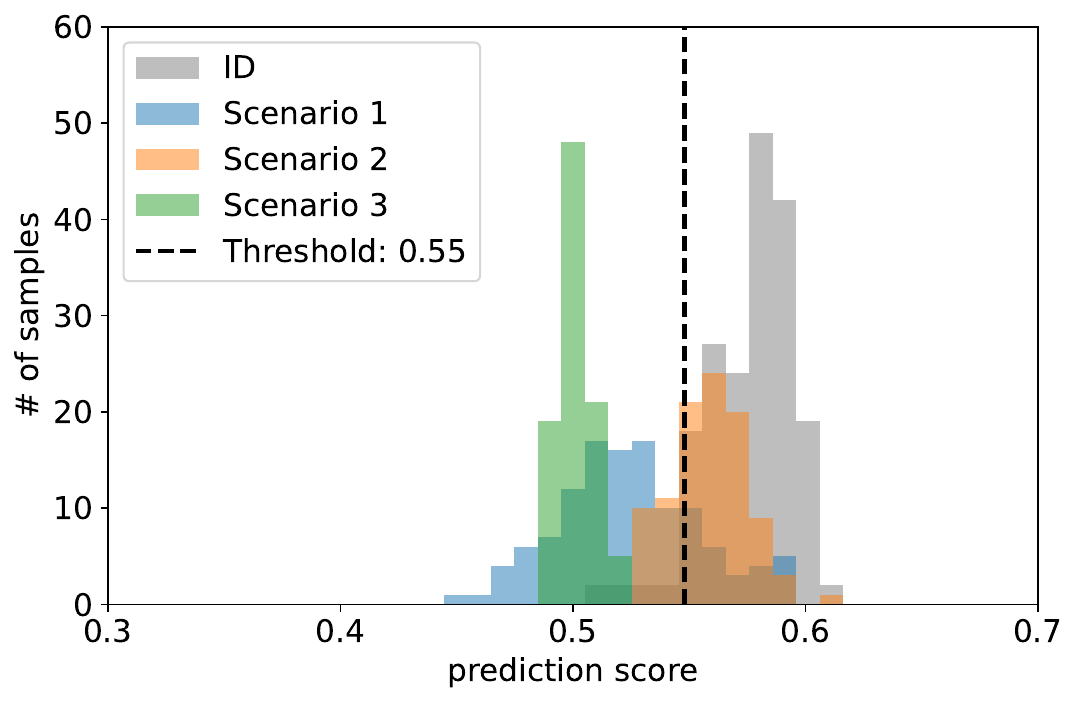}}
\label{fig:bird_contra}
\subfigure[Binary Classification Score.]{
\includegraphics[width=0.32\linewidth]{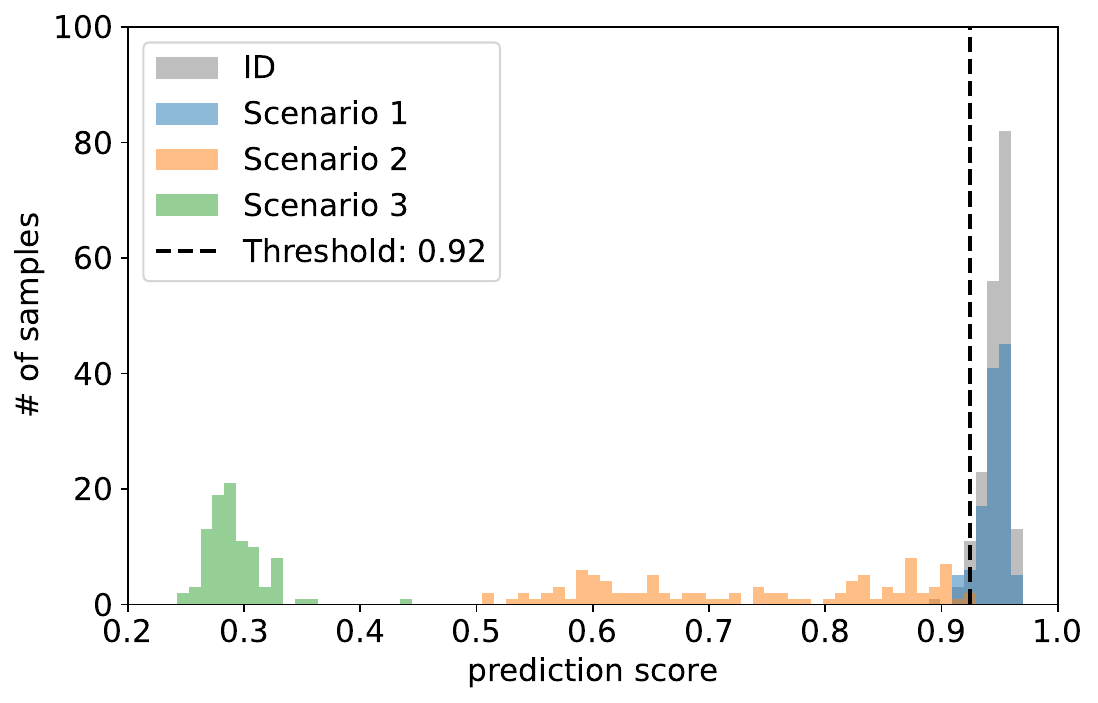}}
\label{fig:bird_class}
\subfigure[Unified OOD Score.]{
\includegraphics[width=0.32\linewidth]{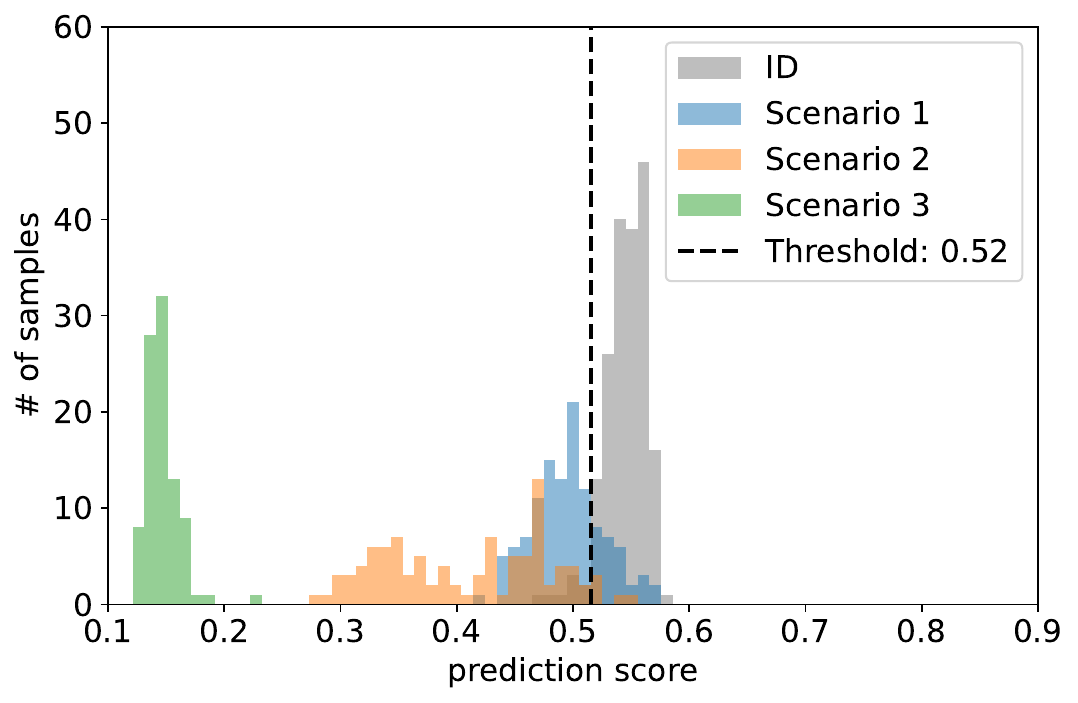}}
\label{fig:bird_score}
\caption{Motivating example of multi-modal OOD detection on CUB-200 dataset. We choose a threshold that includes 95\% of in-distribution test samples~\cite{mingdelving}, and identify a test sample as OOD if its prediction score is below that threshold, i.e., to the left of the dashed line. If an OOD detector works effectively, the prediction scores for all samples in all three OOD scenarios should be concentrated to the left of the dashed line. We can observe from (a) that contrastive learning can only detect OODs in scenarios 1 and 3, but fails to detect scenario 2. Conversely, a binary classifier in (b) can identify OOD samples in scenarios 2 and 3, but it does not work in scenario 1. Motivated by these observations, we develop a general-purpose OOD detection model for multiple OOD scenarios.}
\label{fig:example}
\end{figure*}

The goal of this work is to develop a general-purpose multi-modal OOD detection model that can identify anomalies in various scenarios in a fine-grained manner. We focus on three different OOD scenarios for multi-sensory data: (1) unaligned pairs of data samples, e.g., an in-distribution (ID) image not aligned with its textual information; (2) aligned pairs of data samples collected from a new domain, e.g., aligned images and text from a new environment with a different distribution from training; (3) the presence of noise in data samples, e.g., samples coming from the same environment but with blurry images due to sensor faults. The primary question is: how can we detect OOD samples from all these OOD scenarios simultaneously? Existing OOD detection approaches, such as the CLIP-based methods~\cite{ming2022delving,esmaeilpour2022zero,fort2021exploring} and weakly-supervised classifications~\cite{ji2020multi,wang2021radar}, only focus on one of these three scenarios, thus failing to generalize to all of them. Fig.~\ref{fig:example} illustrates a motivating example of multi-modal OOD detection on CUB-200 dataset~\cite{wah2011caltech} using a CLIP-based method and a weakly-supervised classifier. We can observe from Fig.~\ref{fig:example} (a) that the CLIP-based contrastive learning can only detect OODs in scenarios 1 and 3, but is not effective in scenario 2. This is because in scenario 2 we deliberately choose the test OOD labelled with ``bird" from MS-COCO data~\cite{lin2014microsoft}, which are similar to those ID samples in the training CUB-200, thus making them more difficult for the CLIP-based method. In contrast, a binary classifier can only identify OOD samples in scenarios 2 and 3, but fails in scenario 1, as illustrated in Fig.~\ref{fig:example} (b). This observation motivates us to develop a new OOD detection model that combines both approaches to complement each other.

In this paper, we propose WOOD, a weakly-supervised multi-modal OOD detection model. As illustrated in Fig.~\ref{fig:wood}, the proposed WOOD model consists of two components: a binary classifier for classifying OOD samples and a contrastive learning module for measuring the similarity scores between multiple data modalities. On the contrastive learning side, we adopt Hinge loss to maximize similarity scores of ID samples and minimize those of OOD samples to better distinguish them. On the binary classifier side, we develop a Feature Sparsity Regularizer to better integrate important features from data of multiple modalities. Then a new scoring metric is designed to fuse the prediction results from these two components. Finally, we evaluate the proposed WOOD model on three real-world benchmark datasets. Experimental results demonstrate that our method can simultaneously detect OOD samples under three different scenarios, which significantly outperforms the CLIP-based baselines.

Contributions of this work are summarized as follows.
\begin{itemize}[noitemsep]
    \item We develop a general-purpose multi-modal OOD detection model that can detect anomalies from three different scenarios. 
    \item We design a new scoring metric that combines the prediction results from both the classifier and contrastive learning to improve detection accuracy.
    \item We adopt the Hinge loss in the contrastive objective to better maximize the difference in latent representations of ID and OOD samples.
    \item We show extensive experimental results to demonstrate that the proposed approach is able to achieve very good performance in  detecting OOD under three different scenarios simultaneously.
\end{itemize}

\section{Preliminaries} \label{sec:preliminary}
In this section, we first present the problem of detecting OOD samples for data of multiple modalities and then review contrastive learning for multi-modal OOD detection.

\subsection{Problem Statement}
We consider the problem of detecting multi-modal OOD samples under three different scenarios, as mentioned in Section~\ref{sec:intro}. In this paper, we use vision-language modeling as a running problem for multi-modal OOD detection. With a batch of $N$ pairs of images and texts, along with their labels, denoted by $\{(x_n, t_n), y_n\}_{n=1}^N$, there is a very small number of samples, $K$, that are OOD in the three scenarios above and the remaining $N-K$ pairs are ID samples. The goal of this work is to distinguish the OOD samples from the ID ones using weakly-supervised learning.

\subsection{Contrastive Representation Learning}
Contrastive representation learning aims to encode pairs of data samples into latent representations by making similar samples close to each other and dissimilar ones far apart. One well-known method is the vision-language pre-trained model CLIP, which jointly trains an image encoder and a text encoder to learn the latent representations from text paired with images using zero-shot learning. Specifically, it encodes a pair of image $x_n$ and text $t_n$ into the latent representations $\mathcal{I}(x_n) $ and $\mathcal{T}(t_n)$, respectively, and then adopts cosine similarity to minimize the distance of their representations $\mathcal{I}(x_n)$ and $\mathcal{T}(t_n)$. Due to its excellent performance in learning the latent representations of images and texts, a few CLIP-based methods have been developed to detect multi-modal OOD samples. However, existing methods can only detect one specific scenario where a given image and its textual information come from a new domain with a distribution shift. In the following section, we develop a general-purpose OOD detection framework to identify OODs arising from several different scenarios.

\section{Related Work}
In this seecion, we review and contrast related work on OOD detection in machine learning.

\noindent
\textbf{Single-Modal OOD Detection.} There exists a plethora of works on single-modal OOD detection~\cite{salehiunified} for machine learning. They can be generally categorized into three types: (i) vision OOD detection, (ii) text OOD detection, and (iii) time-series OOD detection~\cite{du2022unknown}. For vision OOD detection, various methods have been developed, including softmax confidence score~\cite{devries2018learning,hein2019relu,hendrycksbaseline,huang2021mos}, energy-based score~\cite{yang2021generalized,sun2021react,sun2022dice}, distance-based method~\cite{sun2022knn,ren2021simple,podolskiy2021revisiting,techapanurak2020hyperparameter,zaeemzadeh2021out,van2020uncertainty}, generative models~\cite{li2022out,xiao2020likelihood}. For instance, Liu et al.~\cite{liu2020energy} proposed an energy-based OOD detection method with theoretical analysis. Sun et al.~\cite{sun2022knn} developed nearest neighbors to improve the flexibility and
generality of OOD detection. For text OOD detection, pre-trained language models~\cite{podolskiy2021revisiting,zhou2021contrastive} are commonly used due to their robustness in identifying OOD samples in natural languages. Other methods, such as data augmentation~\cite{zhan2021out} and contrastive learning~\cite{zhou2022knn,zhou2021contrastive}, have also been developed for OOD detection. Furthermore, some researchers have focused on OOD detection in time-series data~\cite{romero2022outlier,kaur2022codit,georgescu2021anomaly}, where several ML models have been developed for video anomaly detection~\cite{georgescu2021anomaly,wang2018abnormal}. Wang et al.~\cite{wang2018abnormal} combined LSTM with CNN to improve anomaly detection using a spatio-temporal auto-encoder. Li et al.~\cite{li2022context} leveraged generative models to predict middle frames based on past and future frames. However, these methods mainly detect OOD samples using unimodal data, such as images and text. In contrast, we develop a general-purpose model that combines multi-modal data, such as images and textual information, to enhance the performance of OOD detection.

\noindent
\textbf{Multi-Modal OOD Detection.} Some studies \cite{sun2020real, wang2021radar} have adopted multi-modal data to improve the OOD detection accuracy based on deep neural networks (DNNs). Wang et al. \cite{wang2021radar} proposed a multi-modal transformer network that combines Radar and LiDAR data to detect radar ghost targets. Ji et al. \cite{ji2020multi} developed a supervised VAE (SVAE) model that integrates sensor data of multiple modalities to detect an anomalous operation mode of the car. To improve the accuracy of detecting abnormal driving segments, Qiu et al. \cite{qiu2022unsupervised} developed an unsupervised contrastive approach that uses generative adversarial networks to extract latent features from five modalities. More recently, CLIP-based methods~\cite{ming2022delving,esmaeilpour2022zero,fort2021exploring} have been developed to detect OOD samples. Esmaeilpour et al.~\cite{esmaeilpour2022zero} designed a zero-shot OOD detection model based on pre-trained CLIP~\cite{radford2021learning} to detect and generate candidate labels for test images of unknown classes. However, this method heavily relies on a set of candidate labels. To overcome this issue, Ming et al.~\cite{ming2022delving} developed a zero-shot OOD detection method, called Maximum Concept Matching (MCM), based on pre-train CLIP model. While MCM has shown good performance on multi-modal OOD detection, it can only detect visual OOD samples rather than for both an image and its corresponding texture description. Hence, it is not applicable to other OOD scenarios we are exploring in this work, such as scenario 1.

Different from prior works, we develop a general-purpose multi-modal OOD detector that can identify OOD samples arising from three different scenarios in a fine-grained manner. Our proposed method leverages both weakly supervised learning and contrastive learning for improving OOD detection.
\section{Proposed Model}
In this section, we describe the proposed multi-modal OOD detection framework and propose a new scoring metric for detecting OODs under three different scenarios.

\begin{figure*}[!tb]
\begin{center}
 \includegraphics[width=\textwidth]{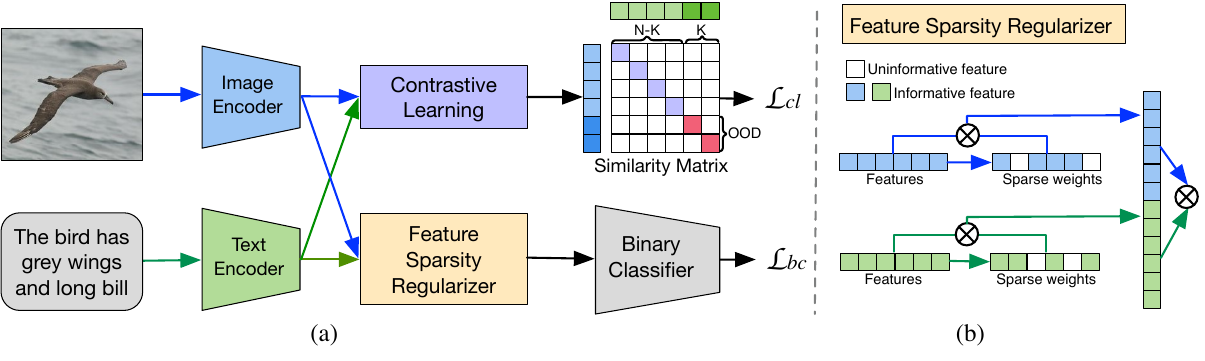}
    \caption{(a) The overall framework of the WOOD detector. It consists of two main parts: 1) a constrastive learning module and 2) a binary classifier. Contrastive learning aims to maximize the difference in similarity scores between ID and OOD samples while the binary classifier is used to predict the probability of being OOD. (b) An expanded detail view of the Feature Sparsity Regularizer in the WOOD framework. This regularizer integrates important features from data of multiple modalities to improve classification accuracy.}
    \label{fig:wood}
\end{center}
\end{figure*}
We propose a weakly-supervised OOD detector, called WOOD, that combines a classifier and contrastive learning to simultaneously detect three different OOD scenarios mentioned in Section~\ref{sec:intro}. Fig.~\ref{fig:wood} illustrates the overall multi-modal OOD detection framework that consists of two components: a contrastive learning module and a binary classifier. The core idea is to use contrastive learning to learn the representations of different data modalities by enforcing the similarity scores of ID pairs to be higher than those of OOD samples. Then WOOD combines the similarity scores from contrastive learning and the prediction results from a binary classifier to identify OOD samples. Below, we detail the two main components in the proposed method.\\

\noindent
\textbf{Contrastive Learning with Hinge Loss}. Inspired by CLIP-based OOD detection methods, we adopt an image encoder and a text encoder to learn the representations of  input pairs (image and text) using contrastive learning. However, unlike existing zero-shot CLIP-based detectors, we add a small number of OOD samples to better separate the representations of ID and OOD samples. Thus, our goal is to maximize the cosine similarity of representations learned from ID samples but minimize those of OOD samples. Let ($x_n$, $t_n$) be a pair of input image and text, and their corresponding representations denote $(\mathcal{I}(x_n), \mathcal{T}(t_n))$. Then, we use the following contrastive loss to minimize the cosine similarity of labeled OOD pairs and maximize that of ID pairs.
\begin{equation}
    \min \mathcal{L}_{1} = -\sum_{n=1}^{N-K} S_{id}(x_n^+,t_n^+) + \sum_{k=1}^{K}  S_{ood}(x_k^-,t_k^-),
\end{equation}

where $S_{id}(x_n^+,t_n^+)=\frac{\mathcal{I}(x_n^+).\mathcal{T}(t_n^+)}{\|\mathcal{I}(x_n^+)\|\|\mathcal{I}(t_n^+)\|}$ and $S_{ood}(x_k^-,t_k^-)=\frac{\mathcal{I}(x_k^-).\mathcal{T}(t_k^-)}{\|\mathcal{I}(x_k^-)\|\|\mathcal{I}(t_k^-)\|}$ represent the cosine similarity between image and text features for ID and OOD pairs, respectively. In addition, $N-K$ and $K$ respectively denote the number of ID and OOD pairs.

In order to further maximize the difference in latent representations between ID and OOD samples, we adopt  Hinge loss to constrain their cosine similarity. In this work, we consider Hinge loss for both ID pairs and labeled OOD pairs in the objective function. 

First, we introduce  Hinge loss for $N-K$ ID samples, as shown in the upper part of the similarity matrix in Fig~\ref{fig:wood}. It is given by
\begin{equation}\footnotesize \label{eq:contrast_id}
    \mathcal{L}_{id} = \sum_{n=1}^{N-K} \left(\frac{1}{N}\sum_{i=1, i \neq n}^N \max \left(0,m - S_{id}(x_n^+, t_n^+) + S_{id}(x_n^+, t_i^-) \right)\right),
\end{equation}
where $m$ is a margin, $S_{id}(x_n^+,t_n^+)$ represents the cosine similarity of $N-K$  aligned ID pairs and $S_{id}(x_n^+,t_i^-)$ represents the cosine similarity between $N-K$ ID images and all $N$ texts (including OOD samples), where each text either does not align with its corresponding ID image or belongs to an OOD sample. In short, the objective, $\mathcal{L}_{id}$, aims to maximize the difference between the aligned ID pairs and incorrect pairings by ensuring that such difference is larger than a margin $m$.

Second, we introduce Hinge loss for $K$ OOD samples for constraining their cosine similarity to a small value. Hence, we have
\begin{equation}\label{eq:contrast_ood}
    \mathcal{L}_{ood} = \sum_{k=1}^{K}\left(\frac{1}{N}\sum_{i=1}^N \max \left(0,-m + S_{ood}(x_k^-, t_i^-)\right)\right),
\end{equation}
where $S_{ood}(x_k^-,t_i^-)$ represents the cosine similarity between each of $K$ OOD images and all $N$ texts (including OOD samples). By definition, the $K$ OOD images should not align with any of the $N-K$ ID texts. Additionally, each OOD image should not be aligned with its corresponding text as described in scenario 1.

By combining the above loss functions, $\mathcal{L}_{id}$ and $\mathcal{L}_{ood}$, for ID and OOD samples, the overall contrastive loss is given by:
\begin{equation}\label{eq:contrast}
    \mathcal{L}_{cl} = \frac{1}{N}\left({L}_{id}+ {L}_{ood}\right).
\end{equation}


\noindent
\textbf{Binary Classifier}. We further adopt weakly-supervised learning to classify OOD samples since recent studies have illustrated that it can significantly outperform unsupervised learning methods by adding some OOD samples~\cite{tian2020few,sultani2018real,majhi2021weakly}. However, the challenge lies in how to integrate image and text features for improved classification accuracy. One naive method is to concatenate their embeddings directly and then feed them into a classifier. But this simple fusion approach does not perform well since the informativeness of different features may vary for different samples. Motivated by prior work~\cite{han2022multimodal}, we develop a feature sparsity regularizer to select and integrate important features from data of the two modalities, as illustrated in Fig.~\ref{fig:wood}(b). Specifically, we train encoder networks $E^{\mathcal{I}}: \mathcal{I}(x_n)\rightarrow w_n^{\mathcal{I}}$ and $E^{\mathcal{T}}: \mathcal{T}(t_n)\rightarrow w_n^{\mathcal{T}}$ to use features obtained from image encoder and text encoders,  These features are updated by a sigmoid activation $\sigma$ to assign higher weights to informative features and lower weights to the uninformative features. Namely, the weight vectors $\sigma(w_n^{\mathcal{I}})$ and $\sigma(w_n^{\mathcal{T}})$ are multiplied with $\mathcal{I}(x_n)$ and $\mathcal{T}(t_n)$ respectively. After that, we fuse the features from two modalities by concatenation, yielding $h_n=\oplus[\mathcal{I}(x_n)\otimes\sigma(w_n^{\mathcal{I}}), \mathcal{T}(t_n)\otimes\sigma(w_n^{\mathcal{T}})]$.
Next, we adopt a binary classifier to identify ID/OOD samples based on the fused features $h_n$. In order to introduce sparsity in the weight vectors, we use $L_1$ normalization $\|\sigma(w_n^{\mathcal{I}})\|_1$ and $\|\sigma(w_n^{\mathcal{T}})\|_1$ and add them to our binary cross entropy (BCE) loss function, which is given by:
\begin{equation} \label{eq:binary_classifier_loss}
    \mathcal{L}_{bc} = \frac{1}{N}\left(\sum_{i=1}^N \text{BCE}(y_i, \hat{y}_i)+ \|\sigma(w_n^{\mathcal{I}})\|_1 + \|\sigma(w_n^{\mathcal{T}})\|_1 \right)
\end{equation}
where $\text{BCE}(y_i, \hat{y_i})=-\left(y_i\log(\hat{y}_i)+(1-y_i)\log(1-\hat{y}_i)\right)$. Note that here $y_i=0$ means the $i$-th test sample is OOD while it is an ID sample when $y_i=1$.\\

\noindent
\textbf{Overall objective}. Finally, we jointly train the contrastive learning and binary classifier for OOD detection. The overall objective for multi-OOD detection is given by
\begin{equation}\label{eq:overall_obj}
    \mathcal{L} = \mathcal{L}_{cl} + \lambda \mathcal{L}_{bc},
\end{equation}
where $\mathcal{L}_{bc}$ is the binary cross-entropy loss for the classifier, and $\lambda$ is the weight for balancing the two terms.

\subsection{New Scoring Metric}
To improve the performance of OOD detection, we introduce a novel scoring metric that combines predictions from the contrastive learning and the binary classifier. The key insight is that we identify an image and text pair as ID only when both the contrastive learning and binary classifier predict that sample as ID. In all other cases, we identify it as OOD.  We codify this condition with the following scoring metric for identifying OOD samples:
\begin{equation}\label{eq:score}
    P_{ood} =1-P_{bc}P_{cl},
\end{equation}
where $P_{bc}$ and $P_{cl}$ denote the prediction results from the binary classifier and contrastive learning, respectively.  Subsequent sections demonstrate that the new scoring metric can help detect OOD samples in all three OOD scenarios effectively.

\subsection{Summary of Proposed Model}
We summarize the proposed WOOD model in Algorithm~\ref{alg:wood}. The basic idea is to map ID and OOD pairs (image and text) into latent representations, and then calculate their cosine similarity. Then, we use Hinge loss to maximize the difference in similarity scores between ID and OOD samples. Moreover, we feed the fused latent representations into a binary classifier to classify ID or OOD samples. Finally, we jointly train the binary classifier and the contrastive learning component for OOD detection.
\begin{algorithm}[!thb]
\caption{The Proposed WOOD Model}
\label{alg:wood}
\begin{algorithmic}[1]
\STATE {\bfseries Input:} A batch of $N$ pairs of images and texts, with $N-K$ ID pairs and $K$ labeled OOD pairs.
\STATE {\bfseries Output:} OOD or ID samples.
\STATE {Encode ID pairs into $\mathcal{I}(x_n^+)$ and $\mathcal{T}(t_n^+)$.} 
\STATE {Encode OOD pairs into $\mathcal{I}(x_n^-)$ and $\mathcal{T}(t_n^-)$.} 
\STATE {Compute Hinge loss on cosine similarity for ID samples in Eq.~\eqref{eq:contrast_id} and OOD samples in Eq.~\eqref{eq:contrast_ood} in contrastive learning.}
\STATE {Compute the total contrastive loss $\mathcal{L}_{cl}$ in Eq.~\eqref{eq:contrast}.}
\STATE {Compute loss of the binary classifier $\mathcal{L}_{bc}$ in Eq.~\eqref{eq:binary_classifier_loss}.}
\STATE {Jointly train the binary classifier and contrastive learning based on the overall objective in Eq.~\eqref{eq:overall_obj}.}
\STATE {Identify OOD samples based on the designed scoring metric in Eq. ~\eqref{eq:score}.}
\end{algorithmic}
\end{algorithm}

\section{Experiments}
In this section, we carry out extensive experimentation to evaluate the performance of the proposed WOOD model on multiple benchmark datasets. Then, we conduct ablation studies to explore how main components in model design and hyperparameters impact OOD detection performance. 

\subsection{Datasets}
We implement experiments on the three real-world datasets: COCO~\cite{lin2014microsoft}, CUB-200~\cite{wah2011caltech}, and MIMIC-CXR~\cite{johnson2019mimic}. COCO and MIMIC-CXR contain images and their corresponding textual descriptions. For CUB-200, the textual information comes from literature~\cite{reed2016learning}.

\noindent
\textbf{Three OOD scenarios}. We generate three different OOD scenarios using the above datasets as follows.
\begin{itemize}[noitemsep,topsep=0.2pt]
    \item \textbf{Scenario 1}. Randomly select a subset of ID images and their textual description from a given dataset and shuffle them so that each image is not aligned with its corresponding textual information. Specifically, select images from one category and the unaligned textual descriptions from another category, ensuring that each pair of OOD image and text are not aligned.  
    \item \textbf{Scenario 2}. Choose OOD samples from another new dataset different from the training data. For instance, when conducting experiments on COCO data, select some OOD pairs of texts and images from Google Conceptual Captions (GCC)~\cite{sharma2018conceptual}.
    \item \textbf{Scenario 3}. Add some Gaussian noise to the ID images such that each image is blurry but its corresponding textual information is correct.
\end{itemize}
Table~\ref{tab:scenario} summarizes the detailed information about generating three OOD scenarios using the above datasets in the experiments.
\begin{table}[htb]
\centering
\caption{Detailed Summary of Our Three OOD Scenarios.}\label{tab:scenario}
\begin{adjustbox}{width=0.6\textwidth}
\begin{tabular}{|l|lll|}
\hline
Scenarios  & \multicolumn{1}{l|}{CUB-200}                                 & \multicolumn{1}{l|}{MIMIC-CXR}                                 & COCO                                 \\ \hline
Scenario 1 & \multicolumn{3}{l|}{\begin{tabular}[c]{@{}l@{}}For sampled pairs of ID images and texts from \\a given dataset, swap their textual descriptions for \\different images, and then label them as OOD.\end{tabular}} \\ \hline
Scenario 2 & \multicolumn{1}{l|}{OOD from COCO-bird}    & \multicolumn{1}{l|}{ROCO~\cite{pelka2018radiology}}                                          &  GCC~\cite{sharma2018conceptual}                                   \\ \hline
Scenario 3 & \multicolumn{3}{l|}{\begin{tabular}[c]{@{}l@{}}For sampled pairs of ID images and texts, add Gaussian \\noise to each image, and then label them as OOD.\end{tabular}}     \\ \hline
\end{tabular}
\end{adjustbox}
\end{table}

\subsection{Model Configurations}
Following prior CLIP-based detectors, we also use the CLIP model (ViT-B/16~\cite{radford2021learning}) as the backbone of the contrastive learning module. The two encoders are $\text{CLIP}_{\text{image}}$ and $\text{CLIP}_{\text{text}}$, which are pre-trained Transformer models for image and text~\cite{radford2021learning} respectively. We do not change the base encoders but fine-tune them with Hinge loss in Eq. (\ref{eq:contrast}) for both feature alignment and OOD detection. Recall that WOOD also has a Feature Sparsity Regularizer module (Figure \ref{fig:wood}), which is a single projection layer (MLP) with a sigmoid activation. We set its hidden size to 512, the same as the dimensions of the output embeddings from $\text{CLIP}_{\text{image}}$ and $\text{CLIP}_{\text{text}}$.  The Binary Classifier is a 3-layer fully connected network with ReLU activation, which outputs a single probability score for binary OOD classification and the layer hidden size is 1024, 512, and 256 respectively. We train the proposed WOOD model using Adam optimizer \cite{kingma2014adam} with learning rate $1e^{-6}$ (following CLIP \cite{radford2021learning}) and stepped learning rate schedule. Additionally, the batch size is set to 128 in all of the experiments and the margin is $m=0.2$ for all the 3 datasets. Regarding the overall training objective in Eq. (\ref{eq:overall_obj}), we choose $\lambda=0.8$ for COCO and CUB-200, and $\lambda=0.2$ for MIMIC-CXR after grid search. Note that subsequent ablation studies in Section~\ref{sec:ablation} explore the impact of these two hyper-parameters on detection performance. During model training, we choose $1\%$ labeled OOD samples for each scenario in order to improve model performance. During inference, we use the same ratio ($25\%$) of test samples for ID and three OOD scenarios. Finally, following previous research~\cite{mingdelving}, we choose a threshold $\delta$ (e.g., 0.6) so that a high fraction of ID data (e.g., 95\%) is above the threshold. Then OOD samples are identified when $P_{ood}<1-\delta$.

\subsection{Baselines}
We compare the WOOD model with these baselines.
\begin{itemize}[noitemsep,topsep=0.2pt]
    \item \textbf{CLIP-BCE}~\cite{liznerski2022exposing}. This model fine-tunes the pre-trained CLIP with a BCE classifier to maximize the similarity of an image and its label for ID and minimize that for OOD samples.
    \item \textbf{MCM-OOD}~\cite{ming2022delving}. This method uses zero-shot CLIP for multi-modal OOD detection based on Maximum Concept Matching. It can only detect OODs in one type of scenario in which a given image is not aligned with its label in the training data.
    \item \textbf{CLIP-Energy}~\cite{liu2020energy}. We adopt the energy score-based CLIP method for OOD detection.
    \item \textbf{WOOD-CL}. In this method, we only use the contrastive learning part of the proposed WOOD model.
    \item \textbf{WOOD-BC}. This approach uses only the binary classifier in the proposed WOOD detector.
\end{itemize}

\begin{table*}[!tb]
\centering
\caption{Performance comparison of different methods for OOD detection on CUB-200 dataset averaged over three random seeds. Higher numbers represent better performance. }\label{tab:cub}
\begin{adjustbox}{width=\textwidth}\small
\begin{tabular}{|c|llll|llll|llll|llll|}
\hline
\multirow{2}{*}{Methods} & \multicolumn{4}{c|}{Scenario 1+ID}                                                    & \multicolumn{4}{c|}{Scenario 2+ID}                                                    & \multicolumn{4}{c|}{Scenario 3+ID}                                                    & \multicolumn{4}{c|}{\textbf{Overall} (ID+OOD)}                                                    \\ \cline{2-17} 
 & \multicolumn{1}{c|}{Accy} & \multicolumn{1}{c|}{Recall} & \multicolumn{1}{c|}{Prec.} & F1 & \multicolumn{1}{c|}{Accy} & \multicolumn{1}{c|}{Recall} & \multicolumn{1}{c|}{Prec.} & F1  & \multicolumn{1}{c|}{Accy} & \multicolumn{1}{c|}{Recall} & \multicolumn{1}{c|}{Prec.} & F1  & \multicolumn{1}{c|}{Accy} & \multicolumn{1}{c|}{Recall} & \multicolumn{1}{c|}{Prec.} & F1  \\ \hline
MCM-OOD  & \multicolumn{1}{l|}{56.3} & \multicolumn{1}{l|}{6.4} & \multicolumn{1}{l|}{56.3} & 11.4 & \multicolumn{1}{l|}{79.2} & \multicolumn{1}{l|}{52.0} & \multicolumn{1}{l|}{86.4} & 64.9 & \multicolumn{1}{l|}{81.0} & \multicolumn{1}{l|}{57.1} & \multicolumn{1}{l|}{87.4} & 69.0 & \multicolumn{1}{l|}{55.6} & \multicolumn{1}{l|}{35.5} & \multicolumn{1}{l|}{93.5} & 51.4 \\ \hline
 CLIP-Energy & \multicolumn{1}{l|}{55.9} & \multicolumn{1}{l|}{5.6} & \multicolumn{1}{l|}{47.2} & 10.0 & \multicolumn{1}{l|}{81.8} & \multicolumn{1}{l|}{59.1} & \multicolumn{1}{l|}{87.8} & 70.6 & \multicolumn{1}{l|}{60.5} & \multicolumn{1}{l|}{1.9} & \multicolumn{1}{l|}{18.8} & 3.5 & \multicolumn{1}{l|}{45.8} & \multicolumn{1}{l|}{20.6} & \multicolumn{1}{l|}{89.3} & 33.5 \\ \hline
  CLIP-BCE & \multicolumn{1}{l|}{55.4} &  \multicolumn{1}{l|}{4.4} & \multicolumn{1}{l|}{41.5} & 7.9 & \multicolumn{1}{l|}{65.4} & \multicolumn{1}{l|}{15.0} & \multicolumn{1}{l|}{64.7} & 24.3 & \multicolumn{1}{l|}{84.5} & \multicolumn{1}{l|}{66.3} & \multicolumn{1}{l|}{89.5} & 76.0 & \multicolumn{1}{l|}{49.5} & \multicolumn{1}{l|}{26.2} & \multicolumn{1}{l|}{91.4} & 40.8 \\ \hline
 WOOD-BC & \multicolumn{1}{l|}{56.6} & \multicolumn{1}{l|}{7.3} & \multicolumn{1}{l|}{52.6} & 30.9 & \multicolumn{1}{l|}{\textbf{96.8}} & \multicolumn{1}{l|}{\textbf{99.5}} & \multicolumn{1}{l|}{\textbf{92.4}} & \textbf{95.9} & \multicolumn{1}{l|}{\textbf{97.0}} & \multicolumn{1}{l|}{\textbf{100}} & \multicolumn{1}{l|}{\textbf{92.5}} & \textbf{96.1} & \multicolumn{1}{l|}{73.9} & \multicolumn{1}{l|}{63.0} & \multicolumn{1}{l|}{96.2} & 75.5 \\ \hline
 WOOD-CL & \multicolumn{1}{l|}{79.6} & \multicolumn{1}{l|}{59.6} & \multicolumn{1}{l|}{90.6} & 71.9 & \multicolumn{1}{l|}{67.2} & \multicolumn{1}{l|}{20.0} & \multicolumn{1}{l|}{69.2} & 53.1 & \multicolumn{1}{l|}{96.9} & \multicolumn{1}{l|}{99.9} & \multicolumn{1}{l|}{92.4} & 96.0 & \multicolumn{1}{l|}{71.7} & \multicolumn{1}{l|}{59.8} & \multicolumn{1}{l|}{96.0} & 73.7 \\ \hline

Ours  & \multicolumn{1}{l|}{\textbf{80.3}} & \multicolumn{1}{l|}{\textbf{61.3}} & \multicolumn{1}{l|}{\textbf{90.8}} & \textbf{73.2} & \multicolumn{1}{l|}{96.2} & \multicolumn{1}{l|}{97.8} & \multicolumn{1}{l|}{92.3} & 95.0 & \multicolumn{1}{l|}{\textbf{97.0}} & \multicolumn{1}{l|}{98.7} & \multicolumn{1}{l|}{92.5} & 96.0 & \multicolumn{1}{l|}{\textbf{86.8}} & \multicolumn{1}{l|}{\textbf{82.5}} & \multicolumn{1}{l|}{\textbf{97.1}} & \textbf{89.2} \\ \hline
\end{tabular}
\end{adjustbox}
\end{table*}

\begin{table*}[!tb]
\centering
\caption{Performance comparison of different methods for OOD detection on MIMIC-CXR dataset averaged over three random seeds. Higher numbers mean better performance.}\label{tab:MIMIC}
\begin{adjustbox}{width=\textwidth}\small
\begin{tabular}{|c|llll|llll|llll|llll|}
\hline
\multirow{2}{*}{Methods} & \multicolumn{4}{c|}{Scenario 1+ID}                                                    & \multicolumn{4}{c|}{Scenario 2+ID}                                                    & \multicolumn{4}{c|}{Scenario 3+ID}                                                    & \multicolumn{4}{c|}{\textbf{Overall} (ID+OOD)}                                                    \\ \cline{2-17} 
 & \multicolumn{1}{c|}{Accy} & \multicolumn{1}{c|}{Recall} & \multicolumn{1}{c|}{Prec.} & F1 & \multicolumn{1}{c|}{Accy} & \multicolumn{1}{c|}{Recall} & \multicolumn{1}{c|}{Prec.} & F1  & \multicolumn{1}{c|}{Accy} & \multicolumn{1}{c|}{Recall} & \multicolumn{1}{c|}{Prec.} & F1  & \multicolumn{1}{c|}{Accy} & \multicolumn{1}{c|}{Recall} & \multicolumn{1}{c|}{Prec.} & F1  \\ \hline

MCM-OOD  & \multicolumn{1}{l|}{55.8} & \multicolumn{1}{l|}{4.6} & \multicolumn{1}{l|}{43.7} & 8.9 & \multicolumn{1}{l|}{64.7} & \multicolumn{1}{l|}{13.4} & \multicolumn{1}{l|}{61.5} & 22.0 & \multicolumn{1}{l|}{65.2} & \multicolumn{1}{l|}{14.6} & \multicolumn{1}{l|}{63.8} & 23.8 & \multicolumn{1}{l|}{39.1} & \multicolumn{1}{l|}{10.4} & \multicolumn{1}{l|}{80.6} & 18.5 \\ \hline
 CLIP-Energy & \multicolumn{1}{l|}{56.6} & \multicolumn{1}{l|}{6.8} & \multicolumn{1}{l|}{51.7} & 12.1 & \multicolumn{1}{l|}{93.2} & \multicolumn{1}{l|}{89.9} & \multicolumn{1}{l|}{91.5} & 90.7 & \multicolumn{1}{l|}{65.4} & \multicolumn{1}{l|}{15.3} & \multicolumn{1}{l|}{91.5} & 24.7 & \multicolumn{1}{l|}{55.0} & \multicolumn{1}{l|}{34.5} & \multicolumn{1}{l|}{93.2} & 50.3 \\ \hline
CLIP-BCE & \multicolumn{1}{l|}{56.7} &  \multicolumn{1}{l|}{7.1} & \multicolumn{1}{l|}{52.3} & 12.5 & \multicolumn{1}{l|}{87.12} & \multicolumn{1}{l|}{73.6} & \multicolumn{1}{l|}{89.9} & 81.0 & \multicolumn{1}{l|}{93.7} & \multicolumn{1}{l|}{95.4} & \multicolumn{1}{l|}{92.0} & 93.7 & \multicolumn{1}{l|}{67.8} & \multicolumn{1}{l|}{53.9} & \multicolumn{1}{l|}{95.6} &  68.9\\ \hline
 WOOD-BC & \multicolumn{1}{l|}{57.8} & \multicolumn{1}{l|}{9.6} & \multicolumn{1}{l|}{60.1} & 16.6 & \multicolumn{1}{l|}{95.7} & \multicolumn{1}{l|}{93.9} & \multicolumn{1}{l|}{\textbf{94.8}} & 94.0 & \multicolumn{1}{l|}{96.9} & \multicolumn{1}{l|}{100} & \multicolumn{1}{l|}{92.3} & 96.0 & \multicolumn{1}{l|}{74.4} & \multicolumn{1}{l|}{63.9} & \multicolumn{1}{l|}{96.2} & 76.8 \\ \hline
 WOOD-CL & \multicolumn{1}{l|}{81.1} & \multicolumn{1}{l|}{63.1} & \multicolumn{1}{l|}{90.9} & 74.4 & \multicolumn{1}{l|}{91.6} & \multicolumn{1}{l|}{85.7} & \multicolumn{1}{l|}{91.1} & 88.2 & \multicolumn{1}{l|}{96.9} & \multicolumn{1}{l|}{99.9} & \multicolumn{1}{l|}{92.3} & 96.0 & \multicolumn{1}{l|}{85.8} & \multicolumn{1}{l|}{81.0} & \multicolumn{1}{l|}{97.0} & 88.4 \\ \hline
Ours  & \multicolumn{1}{l|}{\textbf{81.6}} & \multicolumn{1}{l|}{\textbf{64.4}} & \multicolumn{1}{l|}{\textbf{91.0}} & \textbf{75.3} & \multicolumn{1}{l|}{\textbf{96.6}} & \multicolumn{1}{l|}{\textbf{99.1}} & \multicolumn{1}{l|}{92.3} & \textbf{95.4} & \multicolumn{1}{l|}{\textbf{96.9}} & \multicolumn{1}{l|}{\textbf{100}} & \multicolumn{1}{l|}{\textbf{92.3}} & \textbf{96.0} & \multicolumn{1}{l|}{\textbf{88.8}} & \multicolumn{1}{l|}{\textbf{85.6}} & \multicolumn{1}{l|}{\textbf{97.2}} & \textbf{91.0} \\ \hline
\end{tabular}
\end{adjustbox}
\end{table*}

\begin{table*}[!tb]
\centering
\caption{Performance comparison of different methods for OOD detection on COCO dataset averaged over three random seeds. Higher numbers represent better performance. }\label{tab:coco}
\begin{adjustbox}{width=\textwidth}\small
\begin{tabular}{|c|llll|llll|llll|llll|}
\hline
\multirow{2}{*}{Methods} & \multicolumn{4}{c|}{Scenario 1+ID}                                                    & \multicolumn{4}{c|}{Scenario 2+ID}                                                    & \multicolumn{4}{c|}{Scenario 3+ID}                                                    & \multicolumn{4}{c|}{\textbf{Overall} (ID+OOD)}                                                    \\ \cline{2-17} 
 & \multicolumn{1}{c|}{Accy} & \multicolumn{1}{c|}{Recall} & \multicolumn{1}{c|}{Prec.} & F1 & \multicolumn{1}{c|}{Accy} & \multicolumn{1}{c|}{Recall} & \multicolumn{1}{c|}{Prec.} & F1  & \multicolumn{1}{c|}{Accy} & \multicolumn{1}{c|}{Recall} & \multicolumn{1}{c|}{Prec.} & F1  & \multicolumn{1}{c|}{Accy} & \multicolumn{1}{c|}{Recall} & \multicolumn{1}{c|}{Prec.} & F1  \\ \hline
MCM-OOD  & \multicolumn{1}{l|}{40.2} & \multicolumn{1}{l|}{4.6} & \multicolumn{1}{l|}{56.7} & 8.3 & \multicolumn{1}{l|}{61.4} & \multicolumn{1}{l|}{22.0} & \multicolumn{1}{l|}{77.6} & 34.3 & \multicolumn{1}{l|}{59.5} & \multicolumn{1}{l|}{73.7} & \multicolumn{1}{l|}{17.9} & 28.8 & \multicolumn{1}{l|}{31.6} & \multicolumn{1}{l|}{11.9} & \multicolumn{1}{l|}{88.4} & 20.9 \\ \hline
 CLIP-Energy & \multicolumn{1}{l|}{40.6} & \multicolumn{1}{l|}{5.0} & \multicolumn{1}{l|}{60.4} & 9.2 & \multicolumn{1}{l|}{60.9} & \multicolumn{1}{l|}{20.4} & \multicolumn{1}{l|}{77.4} & 32.3 & \multicolumn{1}{l|}{51.5} & \multicolumn{1}{l|}{0.2} & \multicolumn{1}{l|}{3.5} & 0.4 & \multicolumn{1}{l|}{28.5} & \multicolumn{1}{l|}{7.8} & \multicolumn{1}{l|}{83.3} & 14.2 \\ \hline
  CLIP-BCE & \multicolumn{1}{l|}{40.2} &  \multicolumn{1}{l|}{4.3} & \multicolumn{1}{l|}{56.4} & 7.9 & \multicolumn{1}{l|}{76.4} & \multicolumn{1}{l|}{54.2} & \multicolumn{1}{l|}{85.3} & 61.2 & \multicolumn{1}{l|}{96.7} & \multicolumn{1}{l|}{98.7} & \multicolumn{1}{l|}{94.4} & 96.5 & \multicolumn{1}{l|}{54.7} & \multicolumn{1}{l|}{42.1} & \multicolumn{1}{l|}{96.5} & 58.1 \\ \hline
 WOOD-BC & \multicolumn{1}{l|}{40.7} & \multicolumn{1}{l|}{5.1} & \multicolumn{1}{l|}{60.8} & 9.4 & \multicolumn{1}{l|}{96.8} & \multicolumn{1}{l|}{98.9} & \multicolumn{1}{l|}{94.3} & 97.3 & \multicolumn{1}{l|}{97.3} & \multicolumn{1}{l|}{100} & \multicolumn{1}{l|}{94.4} & 97.1 & \multicolumn{1}{l|}{64.2} & \multicolumn{1}{l|}{54.6} & \multicolumn{1}{l|}{97.2} & 69.9 \\ \hline
 WOOD-CL & \multicolumn{1}{l|}{\textbf{97.0}} & \multicolumn{1}{l|}{\textbf{98.4}} & \multicolumn{1}{l|}{\textbf{96.8}} & \textbf{97.6} & \multicolumn{1}{l|}{93.2} & \multicolumn{1}{l|}{91.0} & \multicolumn{1}{l|}{93.9} & 92.4 & \multicolumn{1}{l|}{97.2} & \multicolumn{1}{l|}{99.9} & \multicolumn{1}{l|}{94.4} & 97.1 & \multicolumn{1}{l|}{96.4} & \multicolumn{1}{l|}{96.8} & \multicolumn{1}{l|}{98.4} & 97.6 \\ \hline
Ours  & \multicolumn{1}{l|}{96.7} & \multicolumn{1}{l|}{97.8} & \multicolumn{1}{l|}{96.8} & 97.3 & \multicolumn{1}{l|}{\textbf{96.8}} & \multicolumn{1}{l|}{\textbf{98.9}} & \multicolumn{1}{l|}{\textbf{94.4}} & \textbf{96.6} & \multicolumn{1}{l|}{\textbf{97.3}} & \multicolumn{1}{l|}{\textbf{100}} & \multicolumn{1}{l|}{\textbf{94.4}} & \textbf{97.1} & \multicolumn{1}{l|}{\textbf{97.8}} & \multicolumn{1}{l|}{\textbf{98.7}} & \multicolumn{1}{l|}{\textbf{98.5}} & \textbf{98.6} \\ \hline
\end{tabular}
\end{adjustbox}
\end{table*}

\subsection{Main Results}
In this subsection, we analyze the experiments to thoroughly evaluate the detection performance of WOOD. Four commonly used metrics - accuracy, recall, precision, and F1 score - are used to measure the prediction results. 

We first evaluate the performance of the proposed WOOD on CUB-200 dataset. Table~\ref{tab:cub} illustrates the comparison results of different OOD detection methods using three random seeds. We can observe from this table that our method is effective in all three OOD scenarios, and its overall performance significantly outperforms the baselines. The reason why WOOD outperforms the baselines is that the binary classifier and the contrastive learning module complement each other for OOD detection. It can be seen that WOOD-CL does not effectively detect OOD in scenario 2 while WOOD-BC can identify OOD with high accuracy. In addition, MCM-OOD does not perform well since it is only designed for detecting visual OOD by querying an image to check whether the returned label from CLIP belongs to training labels. In contrast, our method can detect both visual and textual OODs. Note that the recall for scenario 1 is not very high, since the test samples have very similar images as those in the training data, with the same label ``bird'' but only slightly different textual descriptions. As a result, it is very challenging to identify a pair of OOD samples when an image is only slightly unaligned with its textual information. 

Next, we also show that WOOD can identify OOD samples in multi-modal medical data, MIMIC-CXR. As illustrated in Table~\ref{tab:MIMIC}, it can be observed that the proposed method effectively detects OODs in all three OOD scenarios and its overall performance is better than the baselines. Similar to CUB-200 dataset, the recall of scenario 1 is not very high, since the test images have the same class label ``chest scans'' as those in the training data and only have slightly different textual descriptions. 


\begin{figure*}[!tb]
\centering
\subfigure[Contrastive Learning Score.]{
\includegraphics[width=0.32\linewidth]{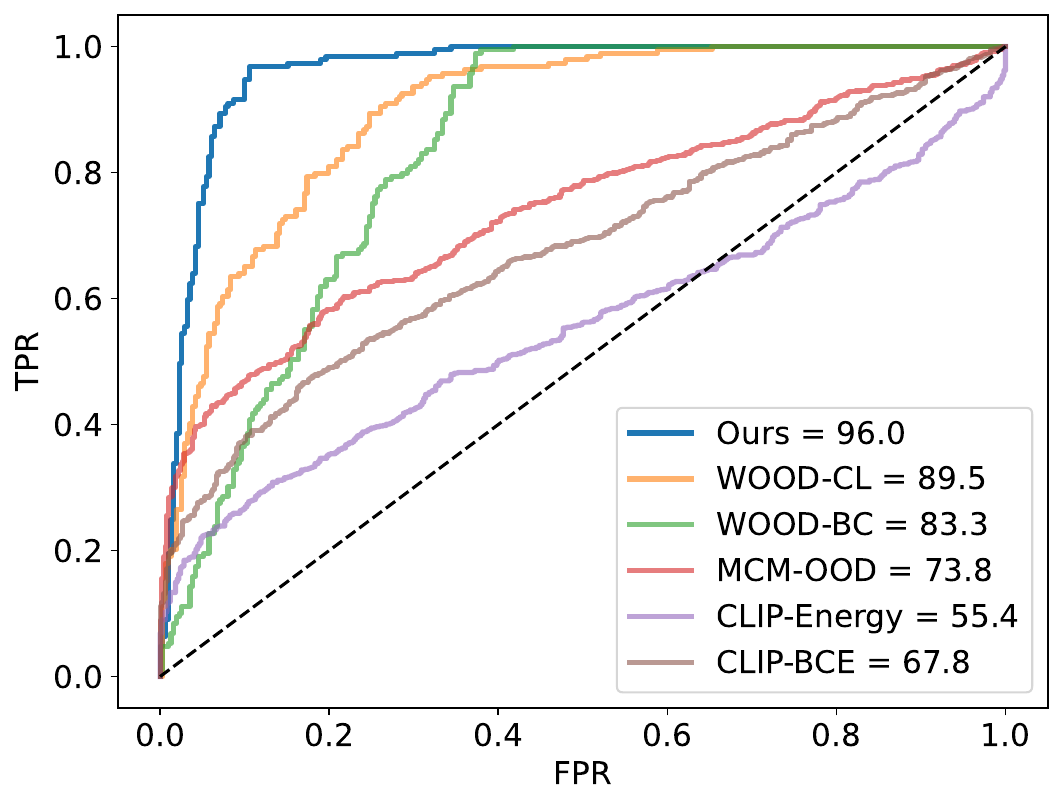}}
\subfigure[Binary Classification Score.]{
\includegraphics[width=0.32\linewidth]{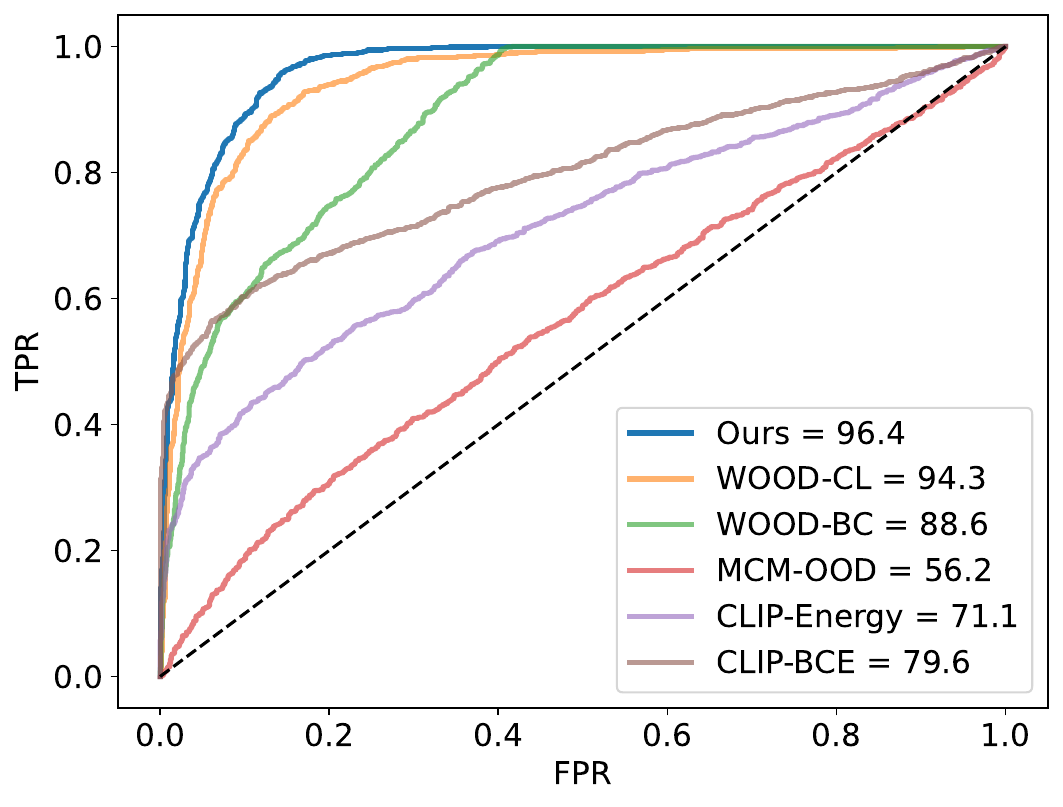}}
\subfigure[Unified OOD Score.]{
\includegraphics[width=0.32\linewidth]{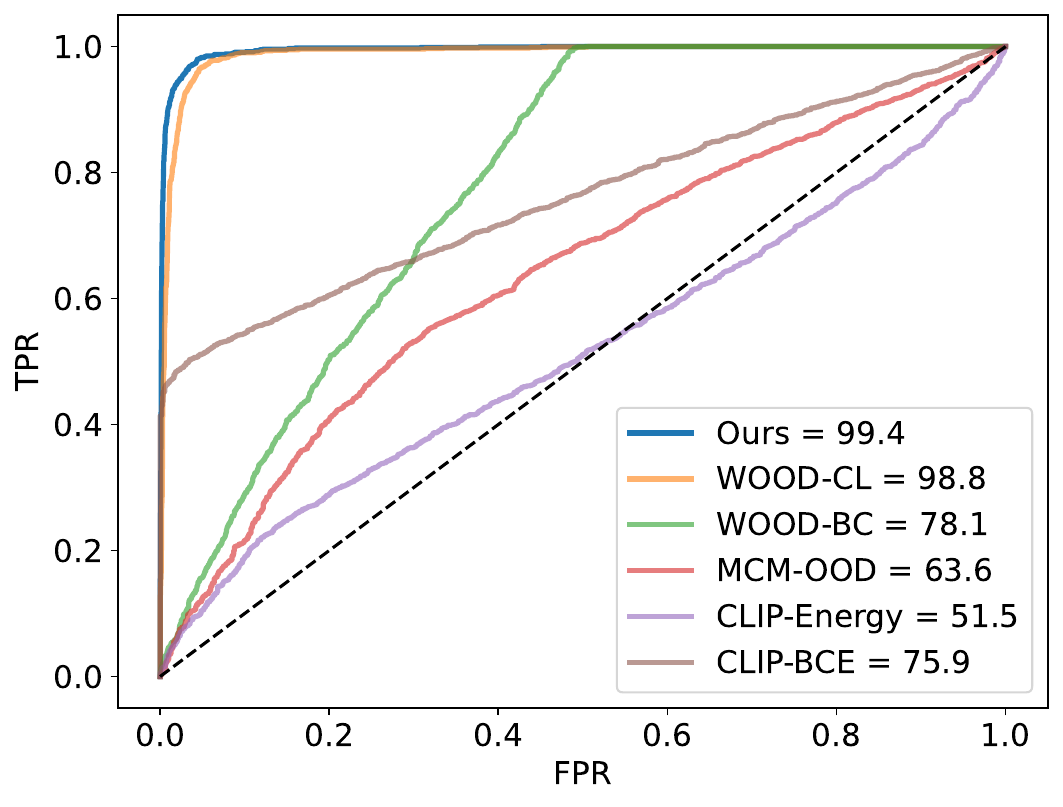}}
\caption{Comparison of AUROC for different methods. We can observe that our method has higher AUROC than the baselines.}\label{fig:auroc}
\end{figure*}

Besides, we apply the proposed WOOD to detect OOD samples on COCO datasets. We compare the detection performance of different methods as shown in Table~\ref{tab:coco}. It can be seen that our method is able to detect OODs in all three OOD scenarios simultaneously while the baseline methods can only detect one or two types of OODs.

Finally, we also compare the AUROC of our method with the baselines on three datasets, as illustrated in Fig.~\ref{fig:auroc}. It can be observed that the proposed WOOD consistently outperforms the baselines in term of AUROC. Based on the above experimental results, we can conclude that WOOD can detect OOD samples arising from different OOD scenarios simultaneously with high accuracy scores.

\subsection{Ablation Study}\label{sec:ablation}
In this section, we conduct ablation studies to investigate the effect of the weight in the objective function, as well as the Hinge loss and its hyperparameter on the performance of OOD detection.

\noindent
\textbf{Effect of Weight in the Objective.} We study the effect of the weight in the objective function ($\lambda$ in Eq. (\ref{eq:overall_obj})) on the performance of OOD detection. We can see from Table~\ref{tab:abs_weight} that the proposed WOOD model has the best performance when $\lambda=0.8$ as it increases from $0.2$ to $1$ on CUB-200 and COCO datasets. Conversely, it performs best when $\lambda=0.2$ on MIMIC-CXR. One possible reason is that the textual description for each image contains about 40 words on average in MIMIC-CXR, which needs more weights in contrastive learning to learn better representations for images and texts.

\noindent
\textbf{Effect of Hinge Loss and its Hyperparameter.} We also explore the impact of Hinge loss (in Eq.~\eqref{eq:contrast_id} and Eq.~\eqref{eq:contrast_ood}) on the detection performance. Table~\ref{tab:abs_margin} shows the detection performance under different margin parameters. When margin $m=0$, it means that the proposed WOOD model does not use Hinge loss. We can observe from Table~\ref{tab:abs_margin} that our method has the best performance when $m=0.2$. In addition, our detection method without the Hinge loss does not perform well (when $m=0$). Therefore, we can conclude that this Hinge loss plays an important role in multi-modal OOD detection. 




\begin{table*}[!tb]
\centering
\caption{Impact of the weight in the objective on detection performance. We report the results averaged over three random seeds. It can be observed that when $\lambda=0.8$, the proposed model has the best performance on CUB-200 and COCO datasets. For MIMIC-CXR data, it performs best as $\lambda=0.2$. }\label{tab:abs_weight}
\begin{tabular}{|c|cccccccccccc|}
\hline
\multicolumn{1}{|l|}{\multirow{2}{*}{}} & \multicolumn{12}{c|}{Dataset}                                                                                                                                                                                                                                                                                                                        \\ \cline{2-13} 
\multicolumn{1}{|l|}{}                  & \multicolumn{4}{c|}{CUB-200}                                                                                     & \multicolumn{4}{c|}{COCO}                                                                                        & \multicolumn{4}{c|}{MIMIC-CXR}                                                                                 \\ \hline
$\lambda$                               & \multicolumn{1}{c|}{Acc.} & \multicolumn{1}{c|}{Recall} & \multicolumn{1}{c|}{Prec.} & \multicolumn{1}{l|}{F1}   & \multicolumn{1}{c|}{Acc.} & \multicolumn{1}{c|}{Recall} & \multicolumn{1}{c|}{Prec.} & \multicolumn{1}{l|}{F1}   & \multicolumn{1}{c|}{Acc.} & \multicolumn{1}{c|}{Recall} & \multicolumn{1}{c|}{Prec.} & \multicolumn{1}{l|}{F1} \\ \hline
0.2                                     & \multicolumn{1}{c|}{85.6} & \multicolumn{1}{c|}{80.4}   & \multicolumn{1}{c|}{96.8}  & \multicolumn{1}{c|}{87.8} & \multicolumn{1}{c|}{94.3} & \multicolumn{1}{c|}{94.1}   & \multicolumn{1}{c|}{98.4}  & \multicolumn{1}{c|}{96.1} & \multicolumn{1}{c|}{88.7} & \multicolumn{1}{c|}{85.4}   & \multicolumn{1}{c|}{97.1}  & 90.9                    \\ \hline
0.4                                     & \multicolumn{1}{c|}{86.4} & \multicolumn{1}{c|}{82.0}   & \multicolumn{1}{c|}{97.1}  & \multicolumn{1}{c|}{88.9} & \multicolumn{1}{c|}{97.0} & \multicolumn{1}{c|}{97.6}   & \multicolumn{1}{c|}{98.5}  & \multicolumn{1}{c|}{98.0} & \multicolumn{1}{c|}{87.9} & \multicolumn{1}{c|}{84.2}   & \multicolumn{1}{c|}{97.1}  & 90.2                    \\ \hline
0.6                                     & \multicolumn{1}{c|}{86.9} & \multicolumn{1}{c|}{82.8}   & \multicolumn{1}{c|}{97.1}  & \multicolumn{1}{c|}{89.4} & \multicolumn{1}{c|}{97.9} & \multicolumn{1}{c|}{98.8}   & \multicolumn{1}{c|}{98.5}  & \multicolumn{1}{c|}{98.7} & \multicolumn{1}{c|}{86.7} & \multicolumn{1}{c|}{82.5}   & \multicolumn{1}{c|}{97.0}  & 89.2                    \\ \hline
0.8                                     & \multicolumn{1}{c|}{87.0} & \multicolumn{1}{c|}{82.9}   & \multicolumn{1}{c|}{97.1}  & \multicolumn{1}{c|}{89.4} & \multicolumn{1}{c|}{97.9} & \multicolumn{1}{c|}{98.8}   & \multicolumn{1}{c|}{98.5}  & \multicolumn{1}{c|}{98.6} & \multicolumn{1}{c|}{85.6} & \multicolumn{1}{c|}{80.7}   & \multicolumn{1}{c|}{97.0}  & 88.1                    \\ \hline
1.0                                     & \multicolumn{1}{c|}{86.8} & \multicolumn{1}{c|}{82.5}   & \multicolumn{1}{c|}{97.1}  & \multicolumn{1}{c|}{89.2} & \multicolumn{1}{c|}{97.8} & \multicolumn{1}{c|}{98.7}   & \multicolumn{1}{c|}{98.5}  & \multicolumn{1}{c|}{98.6} & \multicolumn{1}{c|}{82.7} & \multicolumn{1}{c|}{75.1}   & \multicolumn{1}{c|}{98.5}  & 85.2                    \\ \hline
\end{tabular}
\end{table*}

\begin{table*}[!tb]
\centering
\caption{Impact of Hinge loss in the objective on detection performance. We report the averaged results from three random seeds. It can be seen that the proposed WOOD performs very well when the margin $m=0.2$.}\label{tab:abs_margin}
\begin{tabular}{|c|cccccccccccc|}
\hline
\multicolumn{1}{|l|}{\multirow{2}{*}{}} & \multicolumn{12}{c|}{Dataset}                                                                                                                                                                                                                                                                                                                        \\ \cline{2-13} 
\multicolumn{1}{|l|}{}                  & \multicolumn{4}{c|}{CUB-200}                                                                                     & \multicolumn{4}{c|}{COCO}                                                                                        & \multicolumn{4}{c|}{MIMIC-CXR}                                                                                 \\ \hline
$\mathcal{L}_{cl}$                      & \multicolumn{1}{c|}{Acc.} & \multicolumn{1}{c|}{Recall} & \multicolumn{1}{c|}{Prec.} & \multicolumn{1}{l|}{F1}   & \multicolumn{1}{c|}{Acc.} & \multicolumn{1}{c|}{Recall} & \multicolumn{1}{c|}{Prec.} & \multicolumn{1}{l|}{F1}   & \multicolumn{1}{c|}{Acc.} & \multicolumn{1}{c|}{Recall} & \multicolumn{1}{c|}{Prec.} & \multicolumn{1}{l|}{F1} \\ \hline
$m=0$                                   & \multicolumn{1}{c|}{71.4} & \multicolumn{1}{c|}{58.5}   & \multicolumn{1}{c|}{95.8}  & \multicolumn{1}{c|}{72.5} & \multicolumn{1}{c|}{65.6}     & \multicolumn{1}{c|}{56.8}       & \multicolumn{1}{c|}{97.4}      & \multicolumn{1}{c|}{71.8}     & \multicolumn{1}{c|}{74.5}   & \multicolumn{1}{c|}{64.0}       & \multicolumn{1}{c|}{96.2}     &  76.9                       \\ \hline
$m=0.1$                                 & \multicolumn{1}{c|}{86.2} & \multicolumn{1}{c|}{81.6}   & \multicolumn{1}{c|}{97.1}  & \multicolumn{1}{c|}{88.7} & \multicolumn{1}{c|}{97.5} & \multicolumn{1}{c|}{98.2}   & \multicolumn{1}{c|}{98.5}  & \multicolumn{1}{c|}{98.3} & \multicolumn{1}{c|}{89.3} & \multicolumn{1}{c|}{85.7}   & \multicolumn{1}{c|}{97.2}  & 89.4                    \\ \hline
$m=0.2$                                 & \multicolumn{1}{c|}{86.9} & \multicolumn{1}{c|}{82.8}   & \multicolumn{1}{c|}{97.1}  & \multicolumn{1}{c|}{89.4} & \multicolumn{1}{c|}{97.8} & \multicolumn{1}{c|}{98.7}   & \multicolumn{1}{c|}{98.5}  & \multicolumn{1}{c|}{98.6} & \multicolumn{1}{c|}{88.8} & \multicolumn{1}{c|}{85.4}   & \multicolumn{1}{c|}{97.1}  & 89.2                    \\ \hline
$m=0.3$                                 & \multicolumn{1}{c|}{86.1} & \multicolumn{1}{c|}{81.5}   & \multicolumn{1}{c|}{97.1}  & \multicolumn{1}{c|}{88.6} & \multicolumn{1}{c|}{97.7} & \multicolumn{1}{c|}{98.6}   & \multicolumn{1}{c|}{98.5}  & \multicolumn{1}{c|}{98.5} & \multicolumn{1}{c|}{88.0} & \multicolumn{1}{c|}{84.3}   & \multicolumn{1}{c|}{97.1}  & 88.7                    \\ \hline
$m=0.4$                                 & \multicolumn{1}{c|}{85.3} & \multicolumn{1}{c|}{80.3}   & \multicolumn{1}{c|}{97.0}  & \multicolumn{1}{c|}{87.9} & \multicolumn{1}{c|}{95.9} & \multicolumn{1}{c|}{96.2}   & \multicolumn{1}{c|}{98.4}  & \multicolumn{1}{c|}{97.3} & \multicolumn{1}{c|}{87.1} & \multicolumn{1}{c|}{83.1}   & \multicolumn{1}{c|}{97.1}  & 87.7                    \\ \hline
\end{tabular}
\end{table*}

\section{Conclusions}
In this paper we developed WOOD, a general-purpose multi-modal and weakly-supervised OOD detection framework that combines contrastive learning and a binary classifier in a weakly-supervised fashion. To achieve this, we employ  Hinge loss in contrastive learning to maximize the difference in similarity scores between ID and OOD pairs. In addition, we introduced a Feature Sparsity Regularizer to combine important features from the two data modalities in the binary classifier. We also incorporated a new scoring metric to fuse the prediction results from both components. The evaluation results demonstrated that the integration of the binary classifier and contrastive learning can achieve high accuracy in detecting all three anomaly scenarios.

\bibliographystyle{abbrv}
\bibliography{reference.bib}


\end{document}